\documentclass[runningheads]{llncs}
\usepackage{graphicx}
\begin{document}
\title{Identifying COVID-19 Fake News in Social Media }
%
%
\author{Tathagata Raha\and
Vijayasaradhi Indurthi\and
Aayush Upadhyaya\and
Jeevesh Kataria\and
Pramud Bommakanti\and
Vikram Keswani \and
Vasudeva Varma}
\authorrunning{Raha et al., 2021}
%
\institute{Information Retrieval and Extraction Lab (iREL)\leavevmode \\ International Institute of Information Technology, Hyderabad\leavevmode \\ \leavevmode \  
\email{\leavevmode \\ \{tathagata.raha, vijayasaradhi.i\}@research.iiit.ac.in,\{aayush.upadhyaya, jeevesh.kataria, pramud.bommakanti, vikram.keswani\}@students.iiit.ac.in, vv@iiit.ac.in}\\
}
\maketitle              
\begin{abstract}
The evolution of social media platforms have empowered everyone to access information easily. Social media users can easily share information with the rest of the world. This may sometimes encourage spread of fake news, which can result in undesirable consequences. In this work, we train models which can identify health news related to COVID-19 pandemic as real or fake. Our models achieve a high F1-score of 98.64\%. Our models achieve second place on the leaderboard, tailing the first position with a very narrow margin 0.05\% points. 

\end{abstract}

\section{Introduction}

Fake news is ubiquitous and is impacting all spheres of life. The impact of fake news is more felt especially when the fake news is related to the health of people, specifically relating to the COVID-19 pandemic. Myths, rumours, unsolicited tips and unverified claims and advises related to COVID-19 can sometimes lead to loss of human life. Factually incorrect advises can sometimes create false sense of health and might delay in getting the required medical help often aggravating the condition. Uninformed people can easily become victims of propaganda and has a huge impact on the society.  Needless to say, identifying fake COVID health news is very important as it can save valuable human life.

NLP has made a significant progress in recent times. Transfer learning has been playing an important role in the areas of NLP research. With the introduction of novel architectures  like Transformer, the field of NLP has been revolutionized. We use RoBERTa, a improved variation of BERT for identifying if the COVID health news is fake or real. 

In this task, at first we have used different simple baseline models like naive bayes, linear classifier, boosting, bagging and SVM models to classify a tweet as fake or not. For getting the tweet embeddings, we have used tf-idf and word2vec. As our advanced models, we have experimented with different kinds of transformers models like bert, roberta, electra, etc.

\section{Background}

The task aims at identifying fake news related to COVID-19 in English language. Given a social media post, we need to classify it as a fake or a real categories. The task here was to train machine learning models which can automatically identify posts related to COVID-19 pandemic as fake or real. These posts include posts from various social media platforms like Twitter, Facebook and Instagram. The task deals with these posts in English language, and specifically those posts which are related to the COVID-19 pandemic. For this task, training data has been provided. More details about this dataset has been given in the following sections. Dhoju et. al ~\cite{dhoju2019differences} do a structural analysis and extract relevant features to train models which can classify health news as fake and real. They achieve a high F1-score of 0.96

\section{Related Work}
The study of fake news related to health has not received much attention. With the COVID-19 pandemic, there has been an increased focus in identifying fake health news. We list some of the recent related work here.

Dai et al. ~\cite{dai2020ginger}  constructed a comprehensive repository, FakeHealth, which includes news contents with rich features, news reviews with detailed explanations, social engagements and a user to user social network. They also do exploratory analysis to understand the characteristics of the datasets and analyse useful patterns.
Waszak et al. ~\cite{waszak2018spread} analyze top news shared on the social media to identify leading fake medical information miseducating the society. They curate top health weblinks in the Polish language social media between 2012 and 2017 and provide detailed analysis.


\section{System overview}


In this shared task\cite{patwa2021overview}, we formulate the problem of identifying a social media post as fake or not as a text classification problem. At first, we have implemented a few simple baseline models like linear classifier and boosting models. Then, we use the transformer architecture and fine-tune different pretrained transformer models like Roberta, bert and electra on the COVID-19 training dataset. We do not do explicit preprocessing because we want the model to learn the patterns of input, like the presence of too many hashtags, too many mentions etc. to help identify the fake news.

We use Huggingface's transformers library ~\cite{wolf2019huggingface} for finetuning the pretrained transformer models. 




\section{Dataset}


We use the dataset collected by Patwa et. al ~\cite{patwa2020fighting}. The dataset consisted of tweets and posts related to COVID-19 obtained from different social-media sites like Twitter, Facebook, Instagram and for each tweet there was a label corresponding to a tweet. The labels were as follows:
\begin{enumerate}
\item \textbf{Fake:} This denotes if a post is falsely claimed or fake in nature. 

Example: \textit{Politically Correct Woman (Almost) Uses Pandemic as Excuse Not to Reuse Plastic Bag https://t.co/thF8GuNFPe \#coronavirus \#nashville}
\item \textbf{Real:} This denotes a verified post or a post which is true. 

Example: \it{The CDC currently reports 99031 deaths. In general the discrepancies in death counts between different sources are small and explicable. The death toll stands at roughly 100000 people today.}
\end{enumerate}
Below in Table 1. we can see the distribution of fake and real labels in training set, validation set and test set respectively.
\begin{table}
\center
\begin{tabular}{|l|c|c|c|}
\hline
\bf Split & \bf \#Samples & \bf \#Fake & \bf \#Real\\ 
\hline
Train & 6420 & 3060 & 3360\\
Validation & 2140 & 1020 & 1120\\
Test & 2140 & 1021 & 1120\\
\hline
\hline
\end{tabular}

\caption{Results on validation set for COVID-19 Fake news identification task for English language}
\label{tab:dev_scores_en}
\end{table}
As we can see that the dataset is fully balanced, hence there was no necessity to perform steps to make the dataset balanced. For preprocessing the dataset, we have taken the following measures:
\begin{itemize}
    \item Lowercasing the words
    \item Replacing irrevelant symbols with spaces
    \item Removing stopwords
\end{itemize}
Below in Table 2., we have provided more dataset statistics like the average, maximum and minimum number of words in the posts of training, testing and validation dataset.
\begin{table}
\center
\begin{tabular}{|l|c|c|c|}
\hline
\bf Split & \bf Average & \bf Maximum & \bf Minimum\\ 
\hline
Train & 27.0 & 1456 & 3\\
Validation & 26.8 & 304 & 3\\
Test & 27.5 & 1484 & 4\\
\hline
\hline
\end{tabular}

\caption{Dataset statistics showing the number of words in different splits of the dataset}
\label{tab:dataset_stats}
\end{table}

\section{Baseline models}
We have implemented different simple baseline models on the COVID dataset.

\textbf{Word embeddings:} The first step is to represent each post as a vector.We have chosen two different word embeddings for getting vector representations for our posts and sentences: \textbf{Word2Vec} ~\cite{mikolov2013efficient} and \textbf{tf-idf} ~\cite{10.1145/3232116.3232152}. For the Word2Vec, we find embeddings for each word and take the mean of embeddings of each to get a 300-dimension vector representation for a text.

\textbf{Models:} After getting the word embeddings, we performed experiments with the six following classifiers:
\begin{itemize}
    \item \textbf{Naive Bayes}: This classifier is a probabilistic classifier that uses Bayes Theorem. On the basis of an event that has occurred previously, it calculates the probability of the current event. ~\cite{raschka2017naive}
    \item \textbf{Logistic regression}: Logistic regression is a statistical model that is used to estimate the probability of a response based on predictor variables. ~\cite{Kowsari_2019}
    \item \textbf{Bagging models (Random Forests)}: An ensemble of Decision Trees that uses a tree-like model for predicting the labels. For the final output, it considers the outputs of all the decision trees that it created. ~\cite{markel2020using}
    \item \textbf{Boosting models (XGBoost)}: Boosting is a general ensemble method where at first a lot of weak classifiers are created and then building a strong classifier by building a model from the training data, then creating a second model that attempts to correct the errors from the first model. XGBoost is a decision-tree-based ensemble Machine Learning algorithm that uses a gradient boosting framework. ~\cite{li2018text}
    \item \textbf{Support Vector Machines}: SVM is a non-probabilistic classifier which constructs a set of hyperplanes in a high-dimensional space separating the data into classes. ~\cite{Hassan_2011}
\end{itemize}

\section{Transformer models}
For our more advanced models we explored different transformer models. Vaswani et al.~\cite{vaswani2017attention}  proposed the transformer architecture. They follow the non-recurrent architecture with stacked self-attention and fully connected layers for both the encoder and decoder. Transformer uses concepts like self attention, multi-head attention, positional embeddings, residual connections and masked attention. 
We used the following pre-trained transformer models from HuggingFace repository and fine-tuned it to our classification task:
\begin{itemize}
    \item bert-base-uncased: 12-layer, 768-hidden, 12-heads, 110M parameters. The model has been pretrained on Book Corpus and the Wikipedia data using the Masked Language Model(MLM) and the Next Sentence Prediction(NSP) objectives. ~\cite{devlin2018bert}
    \item distilbert-base-uncased: 6-layer, 768-hidden, 12-heads, 66M parameters. It is a smaller model than BERT which is a lot cheaper and faster to train than BERT. ~\cite{sanh2020distilbert}
    \item roberta-base: 12-layer, 768-hidden, 12-heads, 125M parameters.RoBERTa~\cite{liu2019roberta} is a Robust BERT approach which has been trained on a much more larger dataset and for much larger number of iterations with a larger batch size of 8k. RoBERTa also removes the NSP objective from the pretraining.
    \item google/electra-base: 12-layer, 768-hidden, 12-heads, 110M parameters. ELECTRA models are trained to distinguish "real" input tokens vs "fake" input tokens generated by another neural network, similar to the discriminator of a GAN. \cite{clark2020electra}
    \item xlnet-base-cased: 12-layer, 768-hidden, 12-heads, 110M parameters. It is similar to BERT but it learns bidirectional context alongwith autoregressive formulation. \cite{yang2020xlnet}
\end{itemize}

\section{Experimental Setup}

We combine both the dev and training dataset and then split them into train and validation in the ratio of 90:10. We train on the training split and evaluate on the validation split.


We do not do any explicit pre-processing like removing the mentions or removing the hashtags because we want the model to learn these patterns.

We use Huggingface's transformers library ~\cite{wolf2019huggingface} for all our experiments. 

The primary evaluation metric for the shared task is the F1 score. It is defined as a  is the harmonic mean of the precision and recall. An F1 score reaches its best value at 1 and worst score at 0. In addition, we report the accuracy metric also.

\begin{table}
\center
\begin{tabular}{|l|c|c|}
\hline
\bf Method & \bf Accuracy & \bf F1-score\\ 
\hline
Naive Bayes Model(tf-idf) & 0.887 & 0.885 \\
Linear Classifier(tf-idf) & 0.901 & 0.893 \\
Bagging Model(tf-idf) & 0.926 & 0.921 \\ 
Boosting Model(tf-idf) & 0.914 &  0.913 \\
SVM Model(tf-idf) & \bf 0.941 & \bf 0.941 \\ \hline
Linear Classifier(word2vec) & 0.883 & 0.879 \\
Bagging Model(word2vec) & \bf 0.915 & \bf 0.912 \\ 
Boosting Model(word2vec) & \bf 0.914 & \bf 0.912 \\
SVM Model(word2vec) & 0.909 & 0.905 \\ \hline
bert-base-uncased & 0.962 & 0.960 \\ 
distilbert-base-uncased & 0.957 & 0.955 \\
roberta-base & \bf 0.982 & \bf 0.982 \\ 
electra-base & \bf 0.981 & \bf 0.981 \\
xlnet-base-cased & 0.948 & 0.944 \\
\hline
\hline
\end{tabular}
\caption{Results on validation set for COVID-19 Fake news identification task for English language. The first section denotes the baseline models on tf-idf. Te second section denotes the baseline models on word2vec. The third section refers to the transformers models.}
\label{tab:dev_scores_en}
\end{table}

\begin{table}
\center
\begin{tabular}{|l|c|c|}
\hline
\bf Method & \bf Accuracy & \bf F1-score\\ 
\hline
SVM Model(tf-idf) &  0.939 & 0.938 \\ \hline
Bagging Model(word2vec) &  0.910 & 0.910 \\ 
Boosting Model(word2vec) &  0.927 & 0.926 \\ \hline
roberta-base &  0.9864 &  0.9864 \\ 
electra-base &  0.9827 &  0.9827 \\
\hline
\hline
\end{tabular}
\caption{Results on the official test set for COVID-19 Fake news identification task for English language. The first section denotes the baseline models on tf-idf. Te second section denotes the baseline models on word2vec. The third section refers to the transformers models.}
\label{tab:test_scores_en}
\end{table}

\section{Results}
Table ~\ref{tab:dev_scores_en} shows the results of our model on the validation dataset. We see that the RoBERTa model gives an F1-score of 0.982 with an accuracy of 0.982 on the validation set. Our Electra model achieves an F1-score of 0.981 and an accuracy of 0.981 on the validation set. We submit these two models for final evaluation on the official test set.

Table ~\ref{tab:test_scores_en} shows the official results of our models on the official test set. We see that the RoBERTa model gives an F1-score of 0.9864 with an accuracy of 0.9864 on the official test set.

Our RoBERTa model achieves 2nd position on the official leader board, 0.05 percentage points less than the best F1 score.

Our Electra model achieves an F1-score of 0.9827 with an accuracy of 0.9827 on the official test set, comparable with the top performing models on the leader board



\section{Conclusion}
Identifying fake COVID-19 news is challenging and going forward it would be useful not only to classify if a social media post is fake or not, but also to give interpretation on why the news is fake or not. We would like to explore on the interpretability of the models.

%
%
%
%
\bibliographystyle{splncs04}
\begin{raggedright}
\bibliography{references}
\end{raggedright}
\end{document}